\documentclass{article}

\PassOptionsToPackage{numbers}{natbib}

\usepackage[preprint]{coml_2025}
\usepackage[utf8]{inputenc}
\usepackage[T1]{fontenc}
\usepackage{amsmath, amssymb}
\usepackage{algorithm}
\usepackage{algorithmic}
\usepackage{graphicx}
\usepackage{subcaption}
\usepackage{booktabs}
\usepackage{xcolor}
\usepackage{microtype}
\usepackage{url}
\usepackage{hyperref}
\usepackage{tikz}
\usepackage{amsthm}

\usetikzlibrary{arrows.meta,positioning,fit,calc,shapes.geometric,shapes.multipart,decorations.pathreplacing}

\title{CAFL-L: Constraint-Aware Federated Learning with Lagrangian Dual Optimization for On-Device Language Models}

\author{
Dongqi Zheng \\
Purdue University\\
\texttt{dqzheng1996@gmail.com} \\
\And
Wenjin Fu \\
Carnegie Mellon University \\
\texttt{wenjinf@andrew.cmu.edu}
}

\begin{document}
\maketitle

\begin{abstract}
We introduce Constraint-Aware Federated Learning with Lagrangian Dual Optimization (CAFL-L), a principled extension of FedAvg that explicitly incorporates device-level resource constraints including energy, communication, memory, and thermal budgets. CAFL-L employs Lagrangian dual optimization to dynamically adapt training hyperparameters---freezing depth, local steps, batch size, and communication compression---while preserving training stability through token-budget preservation via gradient accumulation. Experiments on a character-level language model demonstrate that CAFL-L achieves superior constraint satisfaction compared to standard FedAvg (reducing memory usage by 20\% and communication by 95\%) while maintaining competitive validation performance, making it practical for deployment on resource-constrained edge devices.
\end{abstract}

\section{Introduction}
The deployment of language models on edge devices such as smartphones, smartwatches, and AR/VR headsets faces significant challenges due to strict resource constraints. While federated learning ~\cite{mcmahan2017communication} enables collaborative model training without centralizing sensitive data, traditional approaches like FedAvg largely ignore \emph{device/system-level constraints} that are critical for edge deployment.

Existing federated learning methods typically focus on communication efficiency~\cite{li2020federated} or statistical heterogeneity~\cite{karimireddy2020scaffold}, but fail to address the multi-dimensional resource constraints inherent in edge devices: limited energy budgets, restricted communication bandwidth, memory pressure, and thermal throttling. This oversight often leads to training configurations that are infeasible for real-world deployment.

\paragraph{Contributions.}(1) A dual-optimization framework that dynamically adapts training hyperparameters based on resource constraints; (2) a token-budget preservation mechanism that maintains training stability under varying resource allocations; (3) a comprehensive evaluation demonstrating superior constraint satisfaction with competitive accuracy. 

\section{Related Work}
\textbf{Resource-Efficient Federated Learning.} Prior work has explored efficiency along single axes. FedProx / federated systems surveys~\cite{li2020federated} consider heterogeneity but do not enforce \emph{explicit multi-resource} budgets. LAG/FedPAQ~\cite{reisizadeh2020fedpaq} tackle communication reduction, whereas our method simultaneously handles energy, communication, memory, and thermal constraints.

\textbf{Constrained Optimization in ML.} Lagrangian methods are widely used for constrained learning~\cite{cotter2019optimization}. Applying them to FL with \emph{multi-dimensional device constraints} remains underexplored; CAFL-L addresses this gap.

\section{Problem Formulation}
\subsection{Standard Federated Learning}
In standard FL, $N$ clients collaboratively minimize:
\begin{equation}
\label{eq:fl}
\min_{w} F(w) = \sum_{i=1}^N \frac{|\mathcal{D}_i|}{|\mathcal{D}|} \,\ell_i(w),
\end{equation}
where $\mathcal{D}_i$ is client $i$'s local dataset and $\ell_i(w)$ is the local loss.

\subsection{Constraint-Aware Formulation}
Let $\mathcal{B} = (E_b, C_b, M_b, T_b)$ be budgets for energy, communication, memory, and temperature. We formulate:
\begin{equation}
\label{eq:constrained}
\min_{w} F(w) \quad \text{s.t.}\quad u_j \le b_j,\;\; j \in \{E,C,M,T\},
\end{equation}
where $u_j$ denotes measured usage for constraint $j$.

Using Lagrangian duality, we solve:
\begin{equation}
\label{eq:lagrangian}
\min_{w} \max_{\lambda \ge 0}\; \mathcal{L}(w,\lambda)= F(w) + \sum_{j} \lambda_j \max(0,\,u_j-b_j),
\end{equation}
with dual variables $\lambda_j \ge 0$.

\begin{figure}[t]
    \centering
    \includegraphics[width=0.96\textwidth]{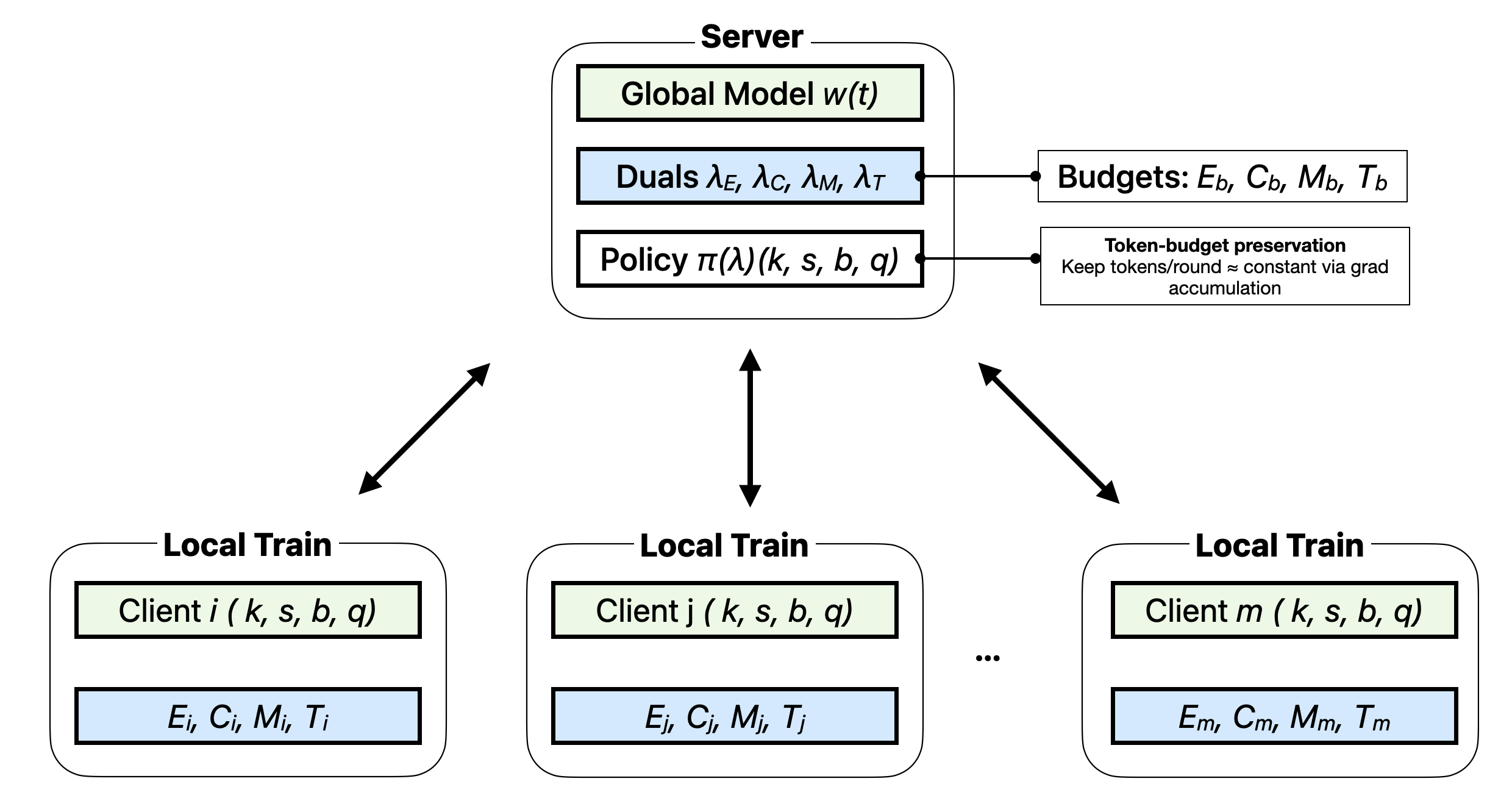}
    \caption{CAFL-L framework overview. The server maintains global model and dual variables. Policy $\pi(\lambda)$ adapts training knobs $(k,s,b,q)$ based on constraint violations. Clients perform local training and report resource usage, which feeds back into dual updates.}
    \label{fig:framework}
\end{figure}

\section{CAFL-L Framework}
The CAFL-L framework is depicted in Fig.~\ref{fig:framework}, with detailed algorithm summarized in Algorithm~\ref{alg:cafl}. Critical components are described follow:
\begin{algorithm}[t]
\caption{CAFL-L Training Algorithm}
\label{alg:cafl}
\begin{algorithmic}[1]
\STATE Initialize global model $w^{(0)}$, dual variables $\lambda_E, \lambda_C, \lambda_M, \lambda_T \leftarrow 0$
\STATE Set baseline parameters: $k_{\text{base}}, s_{\text{base}}, b_{\text{base}}$
\FOR{round $t = 1, 2, \ldots, R$}
    \STATE Evaluate global model: $(acc, loss) \leftarrow \text{evaluate}(w^{(t-1)})$
    \STATE Select client subset $\mathcal{S}_t$ uniformly at random
    \STATE Compute policy: $(k, s, b, q) \leftarrow \pi(\lambda)$
    \STATE Initialize usage accumulators: $U_E, U_C, U_M, U_T \leftarrow 0$
    \FOR{each client $i \in \mathcal{S}_t$}
        \STATE Send global model $w^{(t-1)}$ and hyperparameters $(k, s, b, q)$
        \STATE Client performs local training:
        \STATE \quad $\Delta w_i, (e_i, c_i, m_i, t_i) \leftarrow \text{LocalTrain}(i, w^{(t-1)}, k, s, b, q)$
        \STATE Receive model update $\Delta w_i$ and usage $(e_i, c_i, m_i, t_i)$
        \STATE Update accumulators: $U_E \mathrel{+}= e_i$, $U_C \mathrel{+}= c_i$, $U_M \mathrel{+}= m_i$, $U_T \mathrel{+}= t_i$
    \ENDFOR
    \STATE Aggregate updates: $w^{(t)} \leftarrow w^{(t-1)} + \frac{1}{|\mathcal{S}_t|} \sum_{i \in \mathcal{S}_t} \Delta w_i$
    \STATE Compute average usage: $u_j \leftarrow U_j / |\mathcal{S}_t|$ for $j \in \{E,C,M,T\}$
    \STATE Update dual variables: $\lambda_j \leftarrow \max(0, \lambda_j + \eta \cdot \mathrm{dz}(u_j/b_j))$
\ENDFOR
\end{algorithmic}
\end{algorithm}

\subsection{Dual Variable Updates}
\begin{equation}
\label{eq:dualupdate}
\lambda_j^{(t+1)} = \max\!\left(0,\, \lambda_j^{(t)} + \eta \cdot \mathrm{dz}\!\left(\frac{u_j^{(t)}}{b_j}\right)\right),
\end{equation}
where $\eta$ is the dual learning rate and $\mathrm{dz}(\cdot)$ is a dead-zone function for stability.

\subsection{Policy Function}
The policy $\pi(\lambda)$ maps duals to $(k, s, b, q)$, where $k$ is the number of unfrozen layers; $s$ local steps; $b$ batch size; $q$ compression level (0=32-bit, 1=8-bit, 2=2-bit):
\begin{align}
k &= \max\!\big(1,\, k_{\text{base}} - \lfloor \alpha_k (\lambda_C + \lambda_M + 0.5\lambda_T) \rfloor\big), \\
s &= \max\!\big(10,\, \lfloor s_{\text{base}} (1 - \beta_s (\lambda_E+\lambda_T)) \rfloor\big), \\
b &= \max\!\big(8,\, \lfloor b_{\text{base}} / (1 + \gamma_b (\lambda_T+\lambda_M)) \rfloor\big).
\end{align}

\subsection{Token-Budget Preservation}
Given $T_{\text{target}} = s_{\text{base}} \times b_{\text{base}}$, we keep effective tokens per round roughly constant:
\begin{equation}
\label{eq:gradaccum}
\texttt{grad\_accum} = \max\!\left(1,\, \left\lceil \frac{T_{\text{target}}}{s \cdot b} \right\rceil\right).
\end{equation}

\section{Experiments and Results}
We evaluate CAFL-L on the Tiny Shakespeare dataset~\cite{karpathy2015char} using a GPT-style transformer with 6 layers, 8 attention heads, and 256-dimensional embeddings ($\sim$1.5M parameters). Federated learning is configured with $N=16$ clients, of which 6 participate per round. The budget limit is defined through a proxy resource model representing constrained edge devices such as smartwatches or AR/VR headsets. Reported values are relative units derived from these proxies rather than direct hardware measurements; they can be adapted or re-scaled for specific device profiles in practical deployments. Details of the resource-estimation proxies are provided in Appendix~\ref{app:res-est}.

\begin{figure}[h]
\centering
\begin{subfigure}{0.48\textwidth}
    \includegraphics[width=\linewidth]{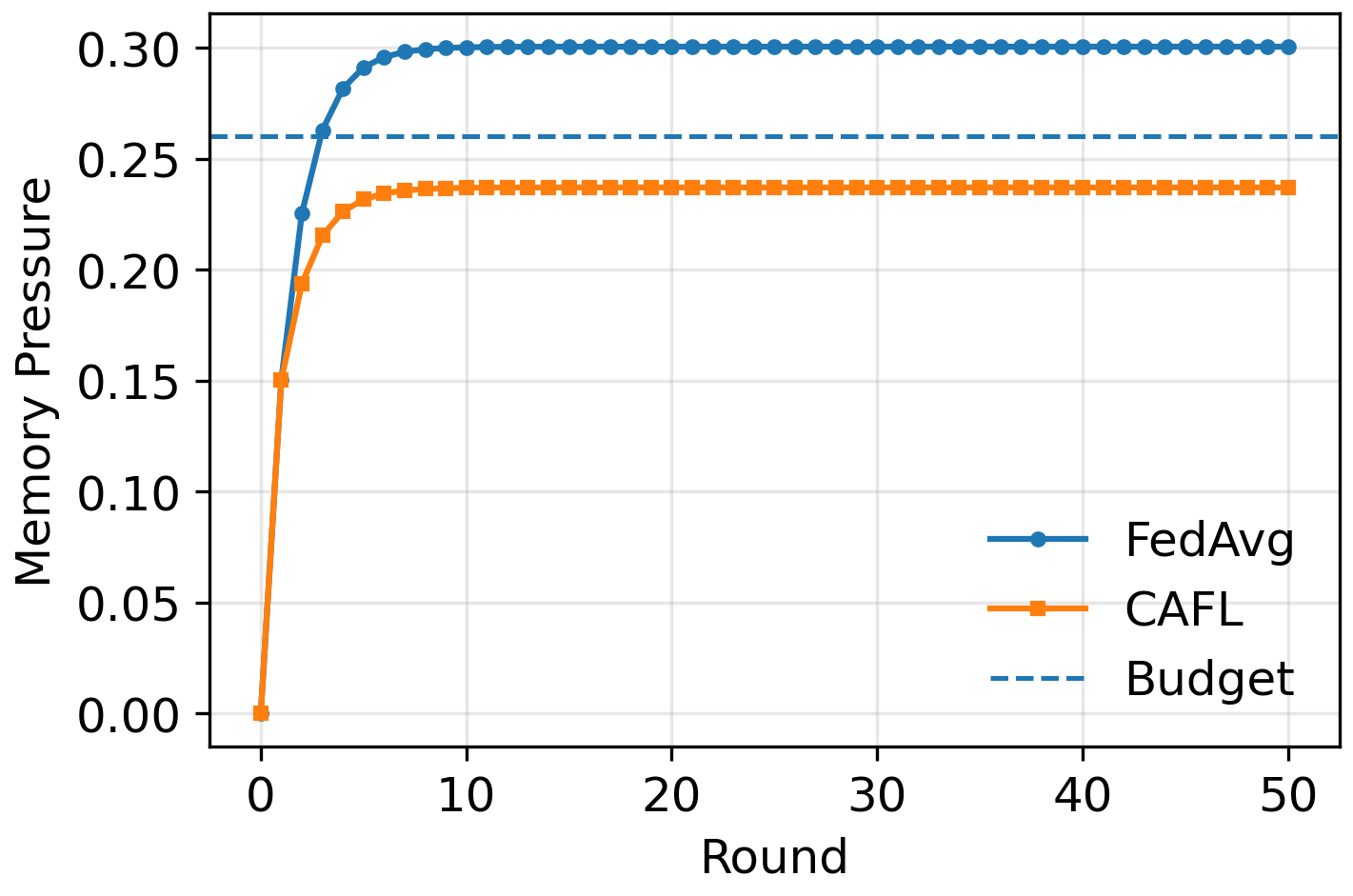}
    \caption{Memory usage}
    \label{fig:mem}
\end{subfigure}
\hfill
\begin{subfigure}{0.48\textwidth}
    \includegraphics[width=\linewidth]{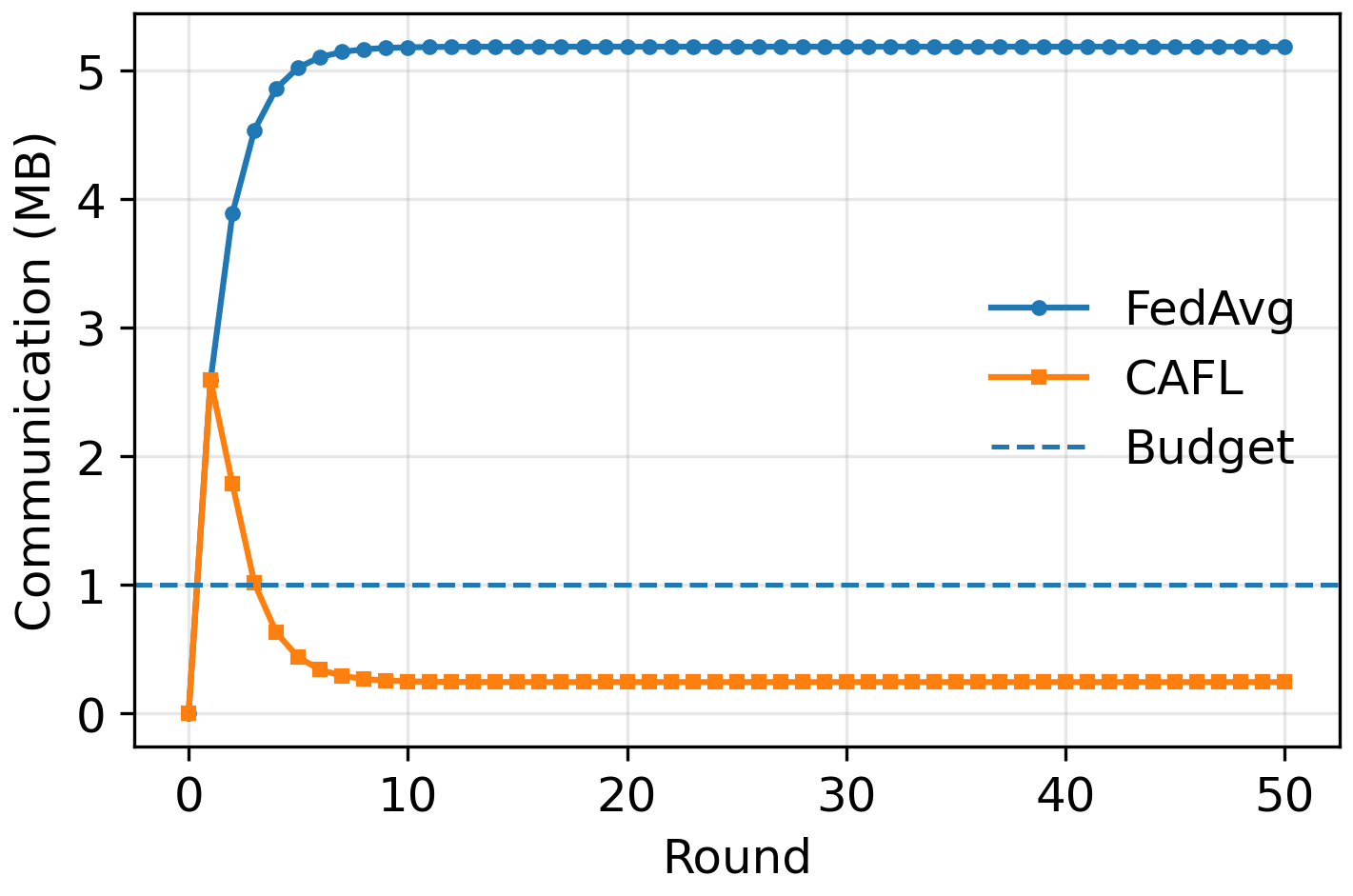}
    \caption{Communication (MB)}
    \label{fig:comm}
\end{subfigure}
\caption{Resource-constraint satisfaction. CAFL-L adaptively manages memory and communication within budgets; while FedAvg keeps violateing them.}
\label{fig:energy_comm}
\end{figure}

\begin{figure}[h]
\centering
\begin{subfigure}{0.48\textwidth}
    \includegraphics[width=\linewidth]{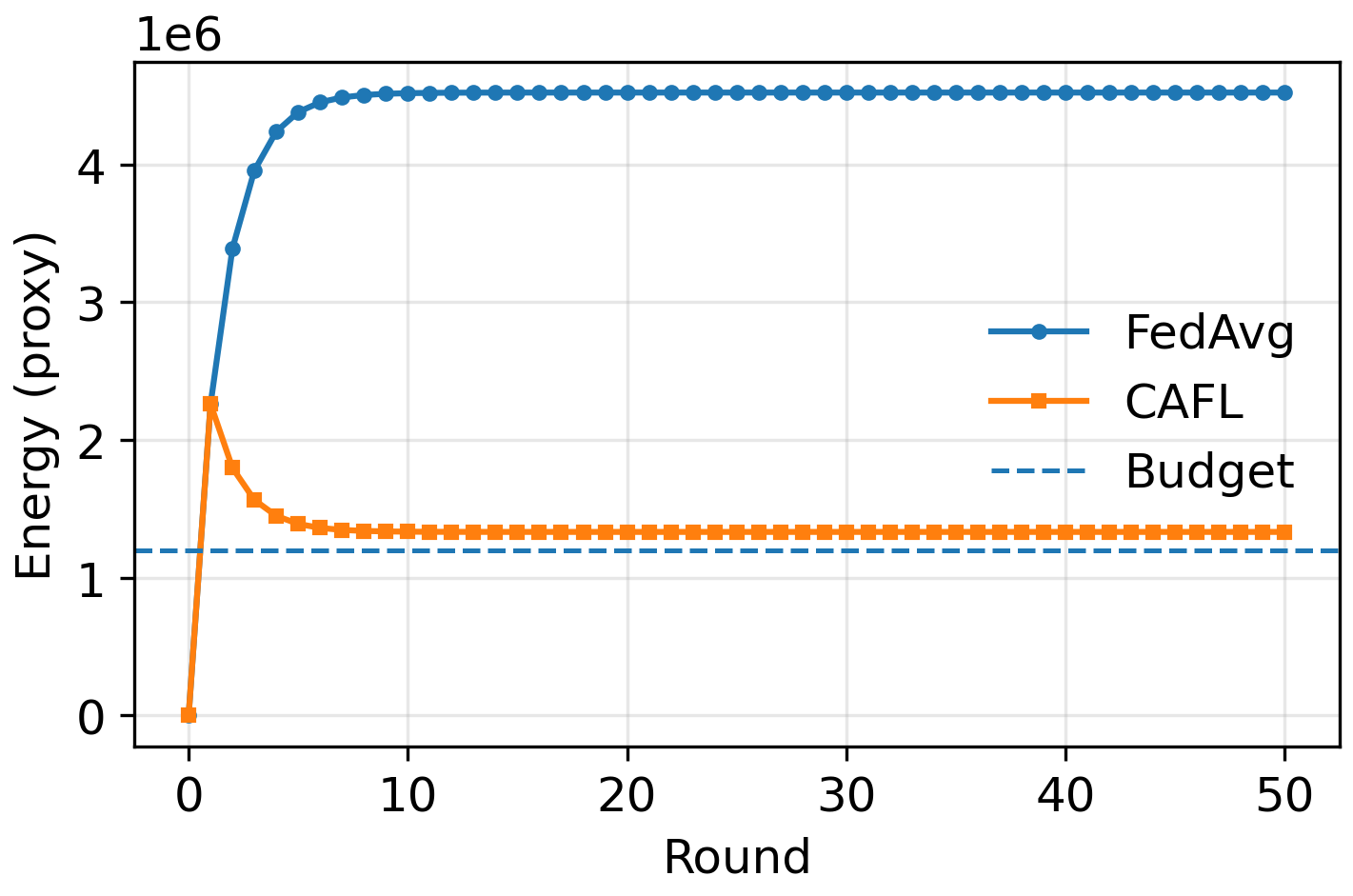}
    \caption{Energy usage}
    \label{fig:energy}
\end{subfigure}
\hfill
\begin{subfigure}{0.48\textwidth}
    \includegraphics[width=\linewidth]{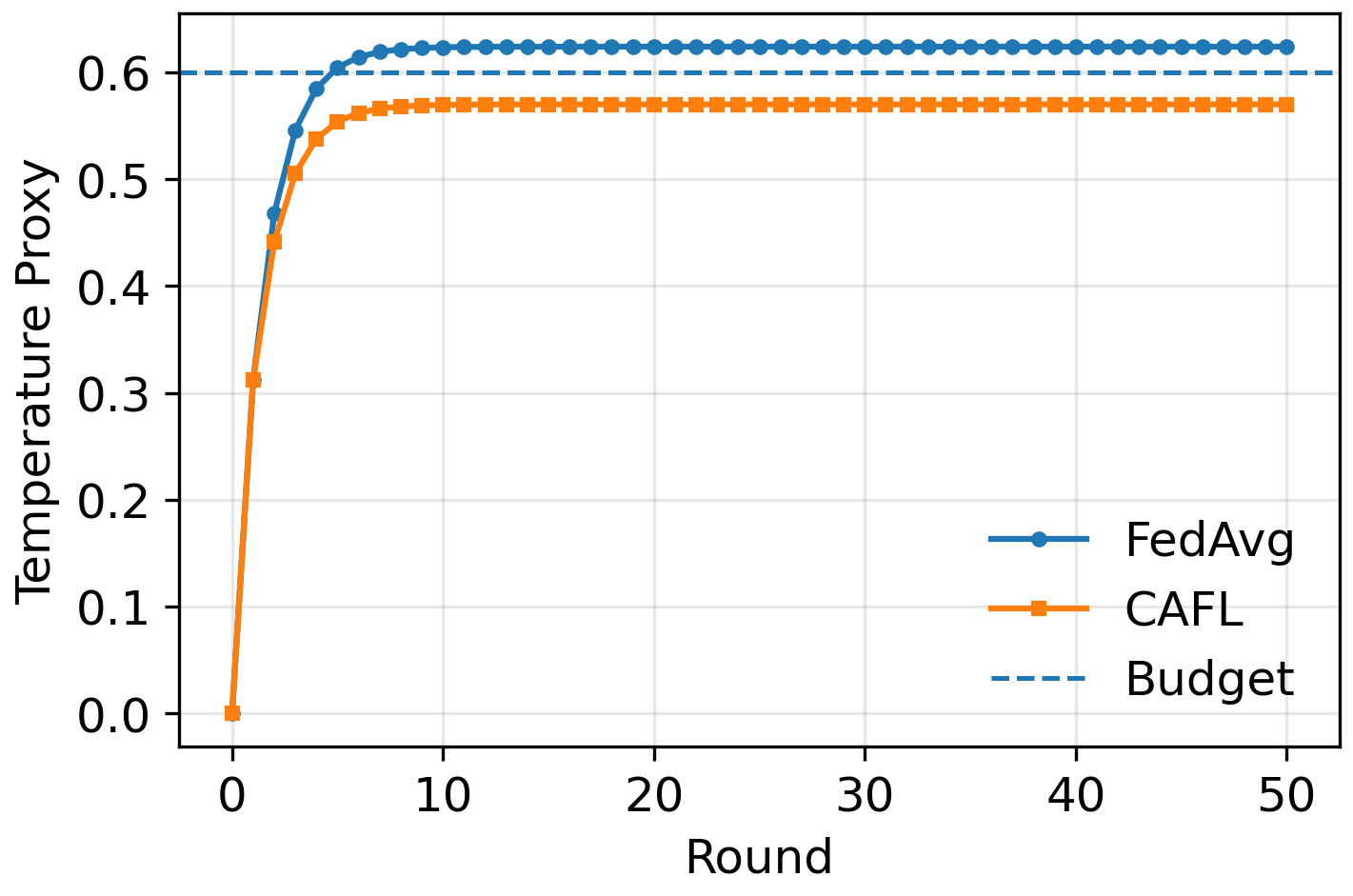}
    \caption{Temperature}
    \label{fig:tempfig}
\end{subfigure}
\caption{Energy and temperature control. CAFL-L prevents energy/thermal runaway by moderating computational intensity and staying near budget.}
\label{fig:energy_temp}
\end{figure}

\begin{figure}[b!]
\centering
\includegraphics[width=0.7\textwidth]{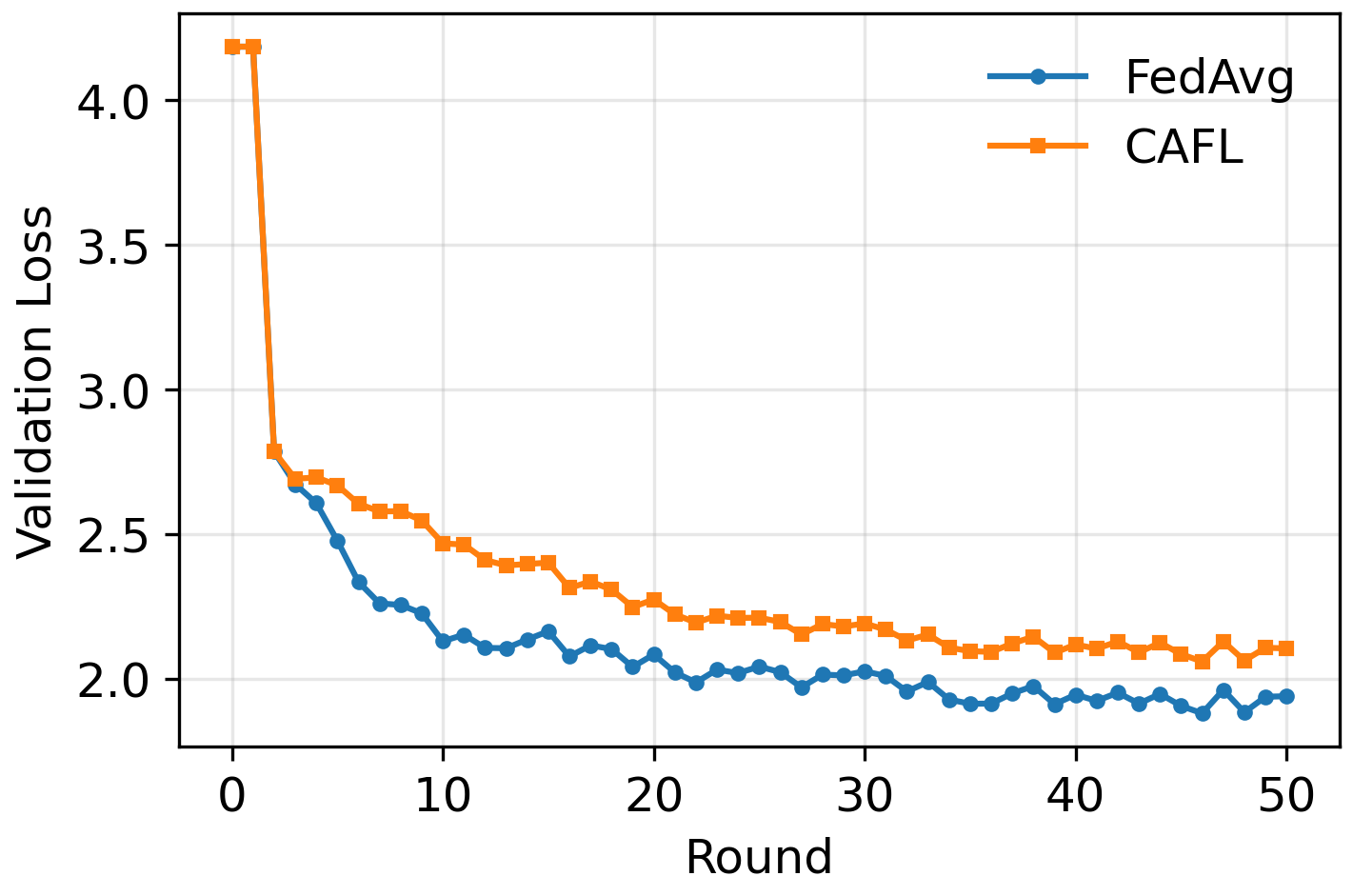}

\caption{CAFL-L shows good convergence (2.10 vs. 1.93), achieving competitive validation loss compared to FedAvg.}
\label{fig:loss.png}
\end{figure}

Figure~\ref{fig:energy_comm} demonstrates CAFL-L's superior constraint satisfaction. While FedAvg consistently violates memory and communication budgets (up to $1.1\times$ and $5.2\times$ the limits), CAFL-L adapts to remain within bounds; by round 50, communication overhead is reduced by roughly 95\% relative to FedAvg. Energy and temperature results (Fig.~\ref{fig:energy_temp}) show CAFL-L stays near or below budget, avoiding thermal issues. CAFL-L also shows promising convergence during training with plots (Fig.~\ref{fig:loss.png}).

Table~\ref{tab:summary} quantifies these results. CAFL-L achieves substantial improvements in resource efficiency: 70\% energy reduction, 95\% communication savings, and 23\% memory improvement, with only a 9\% increase in validation loss, the gap of which can be further reduced by more hyper parameters tuning and training rounds. This is particularly significant for mobile networks where data costs and latency are critical concerns.

\section{Conclusion}
We presented CAFL-L, a constraint-aware federated learning framework that extends FedAvg with principled resource budget enforcement through Lagrangian dual optimization. By dynamically adapting training hyperparameters and preserving effective training tokens, it can achieve superior constraint satisfaction while maintaining competitive validation performance. CAFL-L represents a significant step toward making federated learning practical for deployment on real-world edge devices with strict resource limitations.
\begin{table}[h]
\centering
\caption{Quantitative comparison averaged over final rounds. CAFL-L achieves superior constraint satisfaction with competitive accuracy.}
\label{tab:summary}
\begin{tabular}{lcccccc}
\toprule
Method & Energy ($\times 10^6$)  & Comm (MB) & Temp & Memory & Val. Loss \\
\midrule
Budget Limit & 1.20 & 0.60 & 1.00 & 0.26 & -- \\
FedAvg       & 4.52 & 5.18 & 0.62 & 0.31 & 1.93 \\
CAFL-L       & 1.35 & 0.28 & 0.57 & 0.24 & 2.10 \\
\midrule
Improvement vs. FedAvg & 70\%$\downarrow$ & 95\%$\downarrow$ & 8\%$\downarrow$ & 23\%$\downarrow$ & 9\%$\uparrow$ \\
\bottomrule
\end{tabular}
\end{table}


\appendix

\section{Technical Appendix}
\subsection{Resource Usage Estimation}
\label{app:res-est}
We use lightweight proxies:
\begin{itemize}
\item\textbf{Energy:} $E \approx \alpha_E \cdot \text{params}_{\text{active}} \cdot s \cdot b$. 
\item\textbf{Communication:} $C \approx \text{sparsity} \cdot \text{params}_{\text{active}} \cdot \text{bytes\_per\_param}(q)$. 
\item\textbf{Memory:} $M \approx \alpha_M \cdot (0.2 + \beta_M \cdot \text{params}_{\text{active}} \cdot b)$. 
\item\textbf{Temperature:} $T \approx \alpha_T \cdot (0.35 + \gamma_T s + \delta_T b)$.
\end{itemize}

\end{document}